\documentclass{Interspeech}
\usepackage{colortbl}  

\usepackage{multirow}
\usepackage{multicol}
\usepackage{arydshln}
\usepackage{booktabs}
\usepackage{cite}

\usepackage{xcolor} 

\interspeechcameraready

\title{OpusLM: A Family of Open Unified Speech Language Models}

\author[affiliation={1}]{Jinchuan}{Tian}
\author[affiliation={1}]{William}{Chen}
\author[affiliation={1}]{Yifan}{Peng}
\author[affiliation={1}]{Jiatong}{Shi}
\author[affiliation={1}]{Siddhant}{Arora}
\author[affiliation={1}]{Shikhar}{Bharadwaj}
\author[affiliation={2}]{Takashi}{Maekaku}
\author[affiliation={2}]{Yusuke}{Shinohara}
\author[affiliation={2}]{Keita}{Goto}
\author[affiliation={1}]{Xiang}{Yue}
\author[affiliation={3}]{Huck}{Yang}
\author[affiliation={1}]{Shinji}{Watanabe}

\affiliation{Language Technologies Institute}{Carnegie Mellon University}{USA}
\affiliation{}{LY Corporation}{Japan}
\affiliation{}{NVIDIA Research}{USA}

\email{tianjinchuan@cmu.edu}

\usepackage{comment}
\usepackage{enumitem,amssymb}
\newlist{todolist}{itemize}{2}
\setlist[todolist]{label=$\square$}
\begin{document}

\maketitle

\begin{abstract}
This paper presents \textbf{Op}en \textbf{U}nified \textbf{S}peech \textbf{L}anguage \textbf{M}odels~(OpusLMs), a family of open foundational speech language models (SpeechLMs) up to 7B.
Initialized from decoder-only text language models, the OpusLMs are continuously pre-trained on 213K hours of speech-text pairs and 292B text-only tokens. 
We demonstrate our OpusLMs achieve comparable (or even superior) performance with existing SpeechLMs in speech recognition, speech synthesis, and text-only capabilities.
Technically, this paper articulates our SpeechLM designs on tokenization, multi-stream language models, and multi-stage training strategies. We experimentally demonstrate the importance of model size scaling and the effect of annealing data selection.
The OpusLMs are all built from publicly available materials and are fully transparent models. We release our code, data, checkpoints, and training logs to facilitate open SpeechLM research\footnote{ \url{https://github.com/espnet/espnet/tree/master/egs2/TEMPLATE/speechlm1}}\footnote{ \url{https://huggingface.co/espnet/OpusLM_7B_Anneal}}.

\end{abstract}

\section{Introduction}
Recent advances in large language models (LLMs) have yielded remarkable progress in text processing \cite{gpt4, llama, llm_survey}. This prosperity quickly spread to the speech domain and invoked the popularity of speech language models (SpeechLMs) \cite{slm_survey, slm_survey2, sds_survey}. Specifically, SpeechLMs are sequential models that convert speech-related tasks into sequential modeling problems and then improve the models' capability by scaling data, computing, and parameter scale. The SpeechLM paradigm also simplifies massive multi-tasking \cite{qwen2_audio, voxtlm, uniaudio, speechx, valle-x} and enables spoken language-based human-machine interactions \cite{moshi, vita-1.5, miniomni2, glmvoice, baichuan, vtblender}.
Overall, we believe the SpeechLM paradigm is a promising avenue toward achieving next-generation speech intelligence \cite{speech_intelligence}.

In the past two years, research on SpeechLMs has grown rapidly, which is largely driven by the open research activities \cite{llama, qwen2_audio, voxtlm, uniaudio, moshi, vita-1.5, miniomni2, glmvoice, baichuan}. The gap between open models and closed commercial systems (\textit{i.e.}, GPT-4o \cite{gpt4o}, Gemini series \cite{team2023gemini} and Doubao\footnote{https://team.doubao.com/en/blog/doubao-realtime-voice-model-is-available-upon-release-high-eq-and-iq}) has substantially narrowed due to these activities. This work follows that of prior open SpeechLMs.
In addition to our technical contributions, we adhere to the transparency principles set forth in \cite{owsm3.1, owsm3.2, olmo20242, lacombe-etal-2024-parler-tts} by releasing all code, data, checkpoints, and training logs associated with our pre-trained SpeechLMs. In contrast to prior works that only provide final models emerging from complex development pipelines, we contend that full transparency offers a comprehensive baseline and robust entry points to foster a wider array of research directions in SpeechLMs.

This paper presents \textbf{Op}en \textbf{U}nified \textbf{S}peech \textbf{L}anguage \textbf{M}odels (OpusLMs), a family of open foundational SpeechLMs.
OpusLMs are decoder-only Transformers of a parameter scale spanning from 135M to 7B parameters. They are designed to accept and generate multi-stream discrete tokens in both speech and text modalities.
Built with the ESPnet-SpeechLM toolkit \cite{espnet_speechlm}, these models are initialized from text LLMs and continuously pre-trained on 213K hours of open-available text-speech pairs together with a 292B text-only corpus, for which they become strong in both automatic speech recognition (ASR) and text-to-speech (TTS) while retaining competitive text-only capabilities. 
On the LibriSpeech Test-Clean benchmark \cite{librispeech}, our OpusLMs achieve a word error rate (WER) of 2.3\% and 4.0\% on ASR and TTS, respectively. OpusLM-7B also achieves an MMLU score \cite{mmlu} of 59.0. These results consistently outperform existing SpeechLM baselines. 
Meanwhile, this paper articulates our SpeechLM designs on tokenization, multi-stream language models, and multiple training stages. Additionally, we demonstrate the necessity of adopting a large model size (e.g., 1.7B) for this multi-modal pre-training. We also show that annealing is beneficial although sensitive to data selection.

\vspace{-5pt}
\section{OpusLM}
\vspace{-5pt}
This section introduces the design of the OpusLMs. The tokenization is described in \S\ref{tokenization}. The language model design is in \S\ref{model_design}. We finally describe our training process in \S\ref{training}. 
Note most modular features in our SpeechLM implementation are supported by the existing ESPnet-SpeechLM toolkit \cite{espnet_speechlm}. 
This paper explores the holistic system design of SpeechLM pre-training with these features.

\vspace{-7pt}
\subsection{Tokenization} \label{tokenization}
\vspace{-5pt}
\noindent \textbf{Audio Tokenization:}
Different from prior works using continuous features \cite{qwen2_audio, salmonn}, the audio representations in OpusLMs are all discrete tokens, which contribute to preserving the ease of implementation and input-output consistency. Specifically, our audio tokenization leverages \textit{semantic} tokens and \textit{acoustic} tokens simultaneously. Details about these tokenizers are in \cite{speartts, audiolm, soundstream}. 
As described in \cite{sds_survey}, the semantic tokens and acoustic tokens are strong in speech understanding and generation, respectively. The combination of these two groups of tokens therefore provides features that are sufficiently informative in both directions. 
In our implementation, each frame-level speech representation $\mathbf{x}_t$ contains $N$ tokens, where the first token ${x}_{t, 1}$ is the semantic token while the remaining ${x}_{t, 2}, ..., {x}_{t, N}$ are acoustic tokens (see upper part of Fig.~\ref{fig:delay}). 

\noindent \textbf{Text Tokenization:}
The OpulsLMs are initialized from pre-trained text LLMs so that their text tokenizers are inherited. Since there is only one effective token $x_{t, 1}$ for each text frame\footnote{Although there is no concrete concept \textit{frame} in text processing, we term each timestamp in time-axis as a \textit{frame} for consistency with speech.} $\mathbf{x}_t$, we append extra $N-1$ padding tokens $\emptyset$ to keep compatibility with audio tokenization (Fig.~\ref{fig:delay}). 

\noindent \textbf{Vocabulary:}
OpusLM is built with a joint vocabulary, which consists of text tokens, semantic tokens, and acoustic tokens. Some special tokens are also appended for control flow. 
Non-text embeddings are randomly initialized with the same standard deviation of the pre-trained text embeddings.

\begin{table*}[t]
    \centering
    \caption{Performance Overview of OpusLMs and prior works. OpusLM-1.7B and Opus-7B are initialized from SmolLM2-1.7B and OLMo-2-7B, respectively. Diff: MMLU degradation compared with the initial text LLM.}
    \vspace{-10pt}
    \scalebox{0.60}{
    \begin{tabular}{llr*{4}{>{\columncolor{blue!10}}c}*{2}{>{\columncolor{red!10}}c}*{3}{>{\columncolor{orange!20}}c}*{3}{>{\columncolor{yellow!20}}c}*{2}{>{\columncolor{green!10}}c}}
    \toprule
         & \multicolumn{2}{c}{} & \multicolumn{4}{c}{\cellcolor{blue!10}Open Source} & \multicolumn{2}{c}{\cellcolor{red!10}Data} & \multicolumn{3}{c}{\cellcolor{orange!20}ASR (WER$\downarrow$)} & \multicolumn{3}{c}{\cellcolor{yellow!20}TTS} & \multicolumn{2}{c}{\cellcolor{green!10}TextLM} \\
         \cmidrule(lr){4-7}\cmidrule(lr){8-9}\cmidrule(lr){10-12}\cmidrule(lr){13-15}\cmidrule(lr){16-17}
         No. & Model & Size & Weight & Train & Infer. & Data & Speech & Text & T-Clean & T-Other & FLEURS & WER($\downarrow$) & SPK-SIM($\uparrow$) & MOS($\uparrow$) & MMLU($\uparrow$) & Diff.($\uparrow$) \\ 
         \midrule
         \multicolumn{17}{c}{Single-Task Models} \\
         \midrule
         1 & Whisper-Medium \cite{whisper} & 769M &  \checkmark &  & \checkmark & & 680k hrs. & & 2.8 & 6.5 & \textbf{6.4} &&&& &\\
         2 & OWSM V3.1 \cite{owsm3.1}      & 1B &  \checkmark & \checkmark & \checkmark & \checkmark & 180k hrs. & & \bf{2.4} & \bf{5.0} & 8.4 &&&&& \\ 
         \hdashline
         3 & ChatTTS \cite{chattss2024}    & 280M& \checkmark &&  \checkmark &  & 40k hrs. &&&&& 7.1 & & 3.52 && \\
         4 & CosyVoice \cite{du2024cosyvoice}  & 300M & \checkmark && \checkmark &  & 172k hrs. &&&&& 5.0 & \textbf{0.51} & \bf{4.15} && \\
         5 & Parler-TTS \cite{lacombe-etal-2024-parler-tts} & 2.3B & \checkmark & \checkmark& \checkmark & \checkmark & 45k hrs. &&&&& \bf{4.7} & & 3.83 &&\\
         \hdashline
         6 & LLaMA-3.2 & 1B  & \checkmark  &  & \checkmark & & & 9T  toks. & & & & & && 32.2 &  \\
         7 & Gemma \cite{team2024gemma}     & 2B  & \checkmark  &  & \checkmark & & & 3T  toks. & & && & & & 42.3 &  \\
         8 & SmolLM2 \cite{smollm2}   & 1.7B & \checkmark &  \checkmark & \checkmark &  \checkmark & & 11T toks. & & & & && & 50.7 &  \\
         8 & LLaMA-2 \cite{llama2} & 7B & \checkmark &  & \checkmark &  && 2T toks. &&&&&&&  45.3 & \\ 
         9 & LLaMA-3.1 \cite{llama} & 7B & \checkmark &  & \checkmark &  && 15.6T toks. &&&&&&& \bf{66.7} & \\ 
         10 & OLMo-2 \cite{olmo20242} & 7B & \checkmark & \checkmark & \checkmark & \checkmark && 4T toks. &&&&&&& 63.4 &\\ 
         \midrule
         \multicolumn{17}{c}{Multi-Task Models} \\
         \midrule
         11 & Mini-Omni2 \cite{miniomni2}  & 500M & \checkmark &  & \checkmark & \checkmark & 9k hrs.      & 5M egs.   & 4.8 & 9.8 & &&&&& \\
         12 & VITA-1.5 \cite{vita-1.5}    & 56B  & \checkmark &  & \checkmark &            & 110k hrs.    & 22M egs.  & 3.4 & 7.5 && & & & & \\
         13 & SpiritLM \cite{spiritlm}    & 7B  & \checkmark && \checkmark &  &  569k hrs.&  307B toks. &  6.0 & 11.0 & & 6.7 & & & 36.9 &  -9.3 \\ 
         14 & GLM-4-Voice \cite{glmvoice} & 9B  & \checkmark  &  & \checkmark &            & 497B frms.   & 10.3T toks.& 2.8 & 7.7 & & 5.6 &&& &  \\
         15 & Moshi  \cite{moshi}      & 7B    & \checkmark &  & \checkmark &            & 7M hrs.      & 4T toks.  & 5.7 &      & & 4.7 & & & 49.8 & -4.5 \\
         \hdashline
         16 & \multirow{2}{*}{OpusLM (ours)} & 1.7B & \checkmark & \checkmark & \checkmark & \checkmark & 213K hrs. & 292B toks. & 2.5 & 5.7  & 8.0  & \textbf{4.0} & 0.61 & 4.01 & 46.2 & -4.5 \\
         17 & & 7B & \checkmark & \checkmark & \checkmark & \checkmark & 213K hrs. & 292B toks. & \textbf{2.3} & \bf{5.2} & \textbf{6.1} & {4.6} & \textbf{0.62} & \textbf{4.03} & \textbf{59.0} & \textbf{-2.7} \\
        \bottomrule
    \end{tabular}}
    \label{tab:overall}
    \vspace{-15pt}
\end{table*}

\vspace{-7pt}
\subsection{Model Design} \label{model_design}
\vspace{-3pt}

\noindent \textbf{Multi-Stream Language Model:}
Our OpusLMs are initialized and extended from pre-trained, single-stream text LLMs so that we can leverage the text capability of these models obtained from trillions of text tokens \cite{smollm2, olmo20242}.
As in \S\ref{tokenization}, our OpusLMs learn the multi-stream sequence $\mathbf{X} = [\mathbf{x}_1, ..., \mathbf{x}_T] \in \mathbb{Z}^{T \times N}$, where $T$ is the number of frames while each frame $\mathbf{x}_t$ could be either speech, text or special symbols. This two-dimensional sequence is not compatible with the vanilla single-stream language model and needs extra considerations. Among the several multi-stream language model frameworks \cite{valle, uniaudio, moshi, musicgen}, we proceed with the delay interleave architecture \cite{musicgen}.

\begin{figure}[t]
    \centering
    \includegraphics[width=0.65\linewidth]{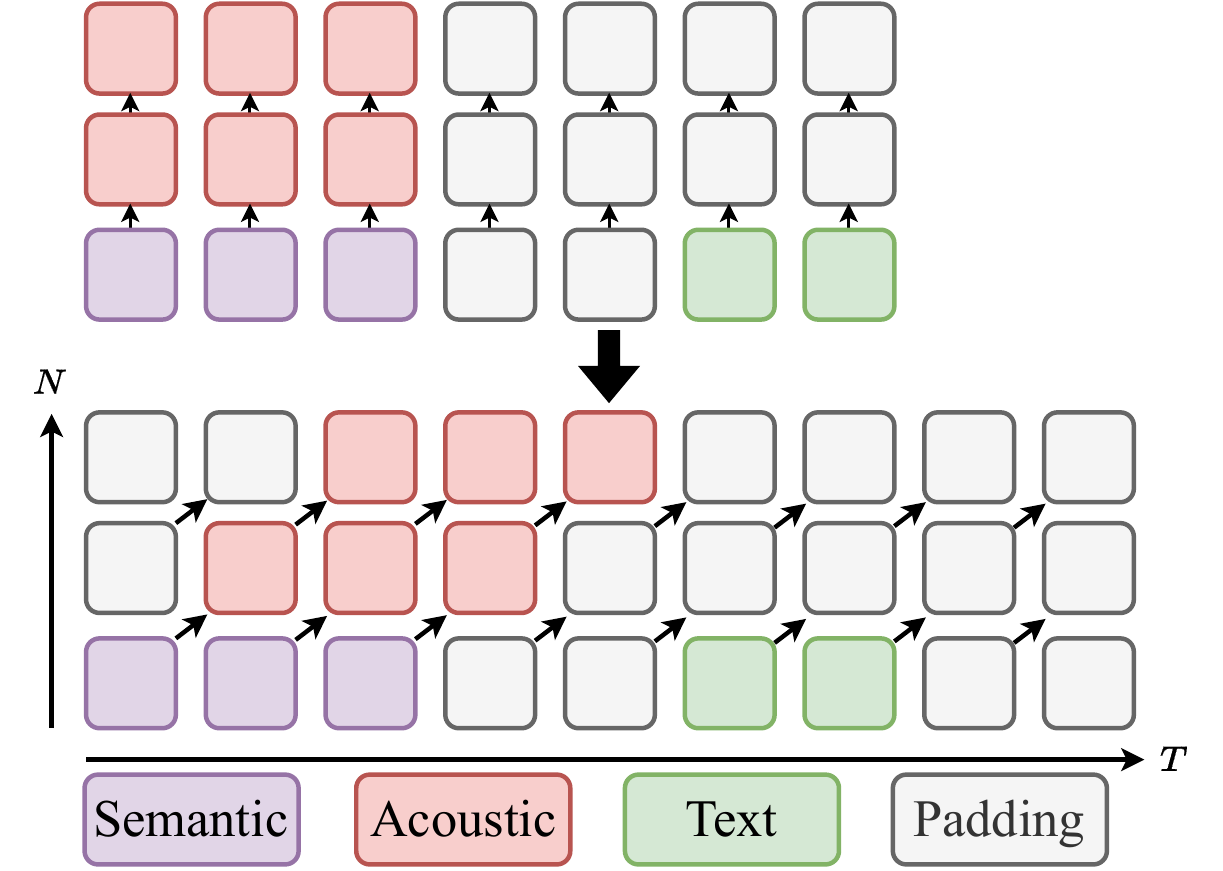}
    \caption{An example of multi-stream speech-text interleaved sequence (upper) and its delay-interleaved sequence (lower). Assume $N=3$, we append extra $N-1$ padding frames to avoid vertical overlap on the speech-text border. Tokens connected by arrows belong to one unique frame.}
    \label{fig:delay}
    \vspace{-15pt}
\end{figure}

\noindent \textbf{Delay Interleave Architecture:}
Instead of modeling on the original $\mathbf{X}$, this architecture works on the delay-interleaved sequence $\mathbf{\hat{X}}$. As shown in Fig.~\ref{fig:delay}, the $\mathbf{\hat{X}}$ is obtained by delaying each token $x_{t,n}$ by $n-1$ frames whenever feasible\footnote{This is feasible only when $t-(n-1) \in [1, T]$. Otherwise $x_{t,n}$ is a padding token $\emptyset$.}:
\begin{equation}
    \hat{x}_{t, n} =x_{t-(n-1), n}.
\end{equation}
Subsequently, the pre-Transformer representation is obtained by summing all embeddings from tokens in each frame:
\begin{equation} \label{embdding}
    \mathbf{h}_t^{\text{pre}} = \sum_{n=1}^{N}\text{Embedding}({\hat{x}}_{t,n}).
\end{equation}
Followed by the forward process of the causal Transformer body to obtain the post-Transformer representation of the next frame:
\begin{equation}
     \mathbf{h}_{t}^{\text{post}} = \text{CausalTransformerBody}(\mathbf{h}_{1}^{\text{pre}}, ..., \mathbf{h}_{t-1}^{\text{pre}}).
\end{equation}
And finally predict all tokens at $\mathbf{\hat{x}}_{t+1}$ parallelly with ${\mathbf{h}}_t$ across each level of $n$:
\begin{equation} \label{eq:bias}
    p(\hat{x}_{t,n}|\mathbf{\hat{X}}_{1:t-1}) = \text{Softmax}(\text{Linear}(\mathbf{h}_t^{\text{post}} + \mathbf{b}_n)).
\end{equation}
where $\mathbf{b}_n$ is a trainable vector to specify which level $n$ in $\mathbf{\hat{x}}_{t+1}$ to predict. 

We adopt this delay interleave architecture for the following reasons. 
First, it is auto-regressive in the $T$-axis, which is aligned with text LLMs.
Second, as shown in Fig.~\ref{fig:delay}, it preserves intra-frame auto-regression\footnote{
Intra-frame auto-regression: for the frame $\mathbf{x}_t$, the prediction of $x_{t, n}$ depends on all $x_{t, < n}$ when predicting $\mathbf{\hat{X}}$ frame-by-frame along $T$-axis. This property is important for speech modeling as there are strong dependencies for speech tokens in the same frame \cite{speartts, moshi}.
}. 
Third, this design still keeps the inference time complexity of $\mathcal{O}(T)$, which is independent of the number of tokens $N$, and then can tolerate a larger $N$.

\noindent \textbf{Task Sequences:}
OpusLMs are trained on four tasks: speech-only, text-only, speech recognition, and speech synthesis. 
The training sequence composition of the former two is single-modal (isolated speech or text sequences).
For ASR, we adopt the spliced \textit{speech-text} sequence. For TTS, we adopt the spliced \textit{text-speech-speech} sequence, where the two \textit{speech} are the speaker prompt and target speech, respectively \cite{valle}.
Different tasks are distinguished by (special) task identifier tokens.

\noindent \textbf{Loss Computing:}
During loss computing, we reweigh the tokens from different sources by the ratio of \textit{text: semantic: acoustic} = 1: 1/2: 1/($N-1$), which is based on our relative importance assumptions that: one text token is equivalent to one speech frame and have summed weight of 1; for all tokens in one speech frame, the semantic token are equivalently important to all acoustic tokens combined (0.5: 0.5). 

As the training sequences consist of several segments, the loss can be applied either to the whole sequences or only the target region of $\mathbf{X}$\footnote{
For ASR and TTS, the target region refers to the text transcription and target speech segmentation, respectively.
}. This option is further discussed in \S\ref{exp:pretraining}


\noindent \textbf{Compatibility with Pre-Trained Text LLM:}
We intentionally ensure our multi-stream language model has the same behavior as the single-stream text LLMs when processing text-only sequences, which satisfy $x_{t, n} = \emptyset, \forall n>1$. Thus, our OpusLMs can be easily exported and coupled with existing text processing tools (e.g., evaluation pipelines, inference engines, etc)\footnote{
In implementation, we force $\text{Embeddings}(\emptyset) = \mathbf{0}$ (in Eq.\ref{embdding}) and $\mathbf{b}_1 = \mathbf{0}$ (in Eq.\ref{eq:bias}). 
}.


\vspace{-7pt}
\subsection{Training Process} \label{training}
\vspace{-3pt}
OpusLMs' training process can be divided into the pre-training stage and the annealing stage. 
This section explains our design concepts, while the detailed specifications are in \S\ref{sec:setup}.

\noindent \textbf{Pre-Training:}
The pre-training stage follows the standard warmup-decay schedule and lasts for 500k updates\footnote{
Due to the computing budget, we only trained the OpusLM-7B for 250k updates, and the model is not fully converged. So this paper mainly discusses the model of 1.7B size.
}. The following findings are noticeable: 
(1) We switch the loss region from the whole sequence to the target region at 250k steps (\S\ref{model_design}). Although we empirically find it beneficial in our small-scale experiments, this benefit is marginal when at scale.
(2) We use a batch size of 1M frames. We also tried the very large batch size (4M frames) following \cite{olmo20242} but found it sub-optimal.
(3) It is essential to keep the text-only corpus in-domain with the text LLM pre-training corpus to avoid severe degradation of text capability.
These three findings are further validated in \S\ref{exp:pretraining}.

\noindent \textbf{Annealing:}
Annealing refers to the practice of decaying the learning rate linearly and quickly to zero in the ending phase of training, which is usually accompanied by a small portion of high-quality data \cite{llama}. Annealing is widely adopted in current text LLM development \cite{llama, olmo20242} and is verified effective in boosting the ultimate model performance.
We show annealing is also effective in SpeechLM training (\S\ref{sec:annealing}). 
With the same pre-trained checkpoint, we conduct two trails of annealing with different data compositions, 85k updates each. Although beneficial in general, we find the ultimate model performance is sensitive to the annealing data selection (\S\ref{sec:annealing}).

Our pre-training speech data is cut off at 30 seconds, for which we observe several length-related problem cases during the evaluation of the pre-trained model. Meanwhile, we expect to fine-tune OpusLMs to downstream applications, some of which may need long-form capability (e.g., multi-turn spoken dialogues). As it is common to append the long-form training phase after normal pre-training in text LLM development \cite{llama}, we also fuse a portion of long-form speech data into the annealing data composition to obtain the long-form capability.

\vspace{-7pt}
\section{Experiments}
\vspace{-3pt}
\subsection{Experimental Setup} \label{sec:setup}
\vspace{-3pt}
\noindent\textbf{Data: }
Our speech data is a mixture of YODAS \cite{yodas}, Emilia \cite{he2024emilia}, and OWSM v3.2 suite \cite{owsm3.2}. We restrict our corpus to English-only and obtain a total volume of 213K hours\footnote{
The YODAS is a subset of the original YODAS v2 generated by the Emilia pipeline \cite{he2024emilia}.
}.
We apply this data mixture to all ASR, TTS, and audio-only tasks, which gives 128B frames\footnote{Around 27k hours of OWSM v3.2 data does not contain speaker information and is not applicable to the TTS task. A portion of these 128B frames are text tokens, i.e., transcriptions.}.
For text corpus, we follow the composition of the pre-trained LLMs \cite{smollm2, olmo20242}.
These datasets are sampled to ensure the text-only data always accounts for 50\% of the training data mixture (292B text tokens in total).

\noindent\textbf{Tokenizers: }
Both semantic and acoustic tokenizers work at 50Hz, and generate 1 and 8 codes per frame respectively (so that $N=9$). We adopt tokenizers that are publicly available for reproducibility. Details are in \cite{espnet_speechlm}

\noindent\textbf{Model: }
For OpusLMs of size 135M, 360M, and 1.7B, we adopt the SmolLM2 series \cite{smollm2} for initialization. These models are trained on up to 11T text tokens. For the OpusLMs of size 7B, we adopt OLMo-2-7B \cite{olmo20242}, which is pre-trained on 4T text tokens. We adopt these models because they are fully transparent and their training text corpus is publicly available.

\noindent\textbf{Training: } 
The model is trained purely in BFloat16 supported by DeepSpeed \cite{deepspeed}, with a context length of up to 8192. We adopt the AdamW optimizer with a peak learning rate of 1e-4 (for OpusLM-7B) or 2e-4 (for others). The warm-up stage lasts 25k steps and then the learning rate decays linearly to 2e-5.
We train these models with up to 64 H100 GPUs, with the Model FLOPs utility (MFU) \cite{mfu} of 28-32\%.

\noindent\textbf{Inference and Evaluation: } We adopt greedy search and top-k sampling for ASR and TTS, respectively. For the top-k sampling, we adopt $k$=30 and temperature=0.7 constantly. By default, we do not conduct any selection over speaker prompts or generated samples before evaluation.
Our text-only inference does not rely on the specific inference method. 
For ASR, we evaluate the word error rate (WER) on the Test-Clean, Test-Other subset of LibriSpeech \cite{librispeech}, and the English subset of FLEURS \cite{FLEURS}. 
For TTS, we evaluate the ASR-WER \cite{whisper}, Speaker Similarity \cite{espnet_spk}, and Proxy MOS \cite{utmos}. TTS evaluation relies on VERSA \cite{shi2024versa}.
For Text-only capability, we evaluate 5-shot MMLU \cite{mmlu} using \texttt{lm\_eval}\footnote{https://github.com/EleutherAI/lm-evaluation-harness}.

\noindent\textbf{Annealing:}
Annealing starts from the learning rate of 5e-5. We have two data compositions for annealing: \texttt{Opt-A}: LibriSpeech \cite{librispeech} + FLEURS \cite{FLEURS} + YODAS \cite{yodas}; \texttt{Opt-B}: LibriTTS \cite{libritts} + VCTK \cite{vctk} + YODAS \cite{yodas}. We use the same YODAS data as in pre-training, but splice the utterances to up to 2 minutes for long-form training. We believe the other four datasets are of high quality due to their rigorous curation\footnote{We upsample these four datasets by 10 times, to balance with the massive YODAS data.}. 
For in-domain evaluation, we adopt the LibriSpeech Test-Other and Test-Clean subset for ASR and TTS, respectively\footnote{Note LibriTTS is roughly a re-segmented version of LibriSpeech so this evaluation should still be considered in-domain for \texttt{Opt-B}.}. 
To verify that the benefit of annealing is general rather than domain-adapted to annealing data, we additionally use the test set of GigaSpeech \cite{gigaspeech} for out-domain evaluation. We also splice the test set of LibriHeavy \cite{kang2023libriheavy} to up to 1 min for long-form evaluation. 

\begin{figure}[t]
    \centering
    \includegraphics[width=\linewidth]{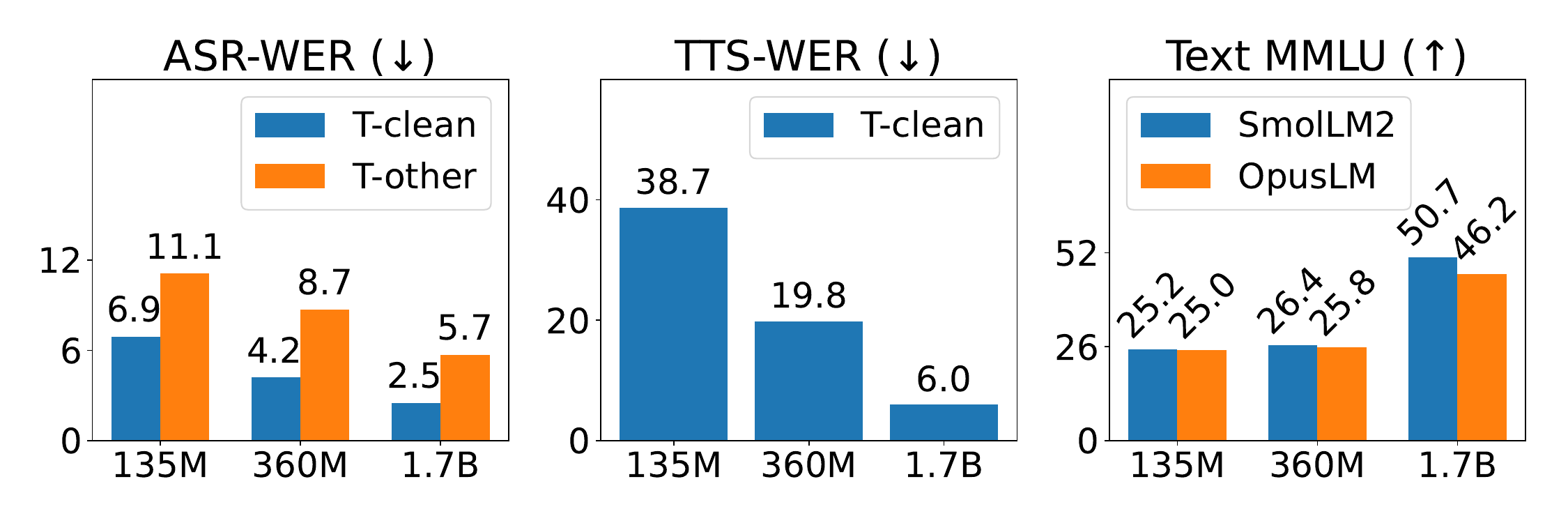}
    \vspace{-20pt}
    \caption{Scaling property of OpusLM on 135M, 360M and 1.7B over ASR, TTS and text-only task}
    \label{fig:scaling}
    \vspace{-15pt}
\end{figure}

\subsection{Overall Results}
In Tab.~\ref{tab:overall}, we compare the performance of our OpusLMs (No. 16-17) with prior works, including single-task ASR models (No.1-2), TTS models (No. 3-5), and text LLMs (No. 6-10) as well as the end-to-end SpeechLMs with both native understanding and generation capabilities \cite{sds_survey} (No. 11-15).

\noindent \textbf{Data and Open Source Practice:}
We follow the prior works to make our OpusLMs fully transparent \cite{lacombe-etal-2024-parler-tts, olmo20242, owsm3.1}, with all code, data, checkpoints, and training logs released. 
Additionally, although our data volume is much smaller than prior works with proprietary data sources (e.g., No.13-15), we demonstrate our public data is still sufficient to achieve competitive results. 

\noindent \textbf{ASR and TTS}:
For ASR, both the 1.7B and 7B versions of OpusLMs outperform other SpeechLMs, especially on the \textit{test-other} subset. We also observe that our models can match the performance of whisper and OWSM models (No. 1-2), which are uniquely designed for speech-to-text tasks like ASR. 
For TTS, our massive pre-trained OpusLM-1.7B outperforms all other competitors (No. 3-5, 13-15) in terms of the WER and achieves comparable results on speaker similarity and proxy MOS score\footnote{The OpusLM-7B has a high WER since it is an intermediate checkpoint. We empirically find the TTS converges slower than ASR.}. 

\noindent \textbf{Text Capability:}
Our OpusLM-7B outperforms all other SpeechLMs on the MMLU metric with a clear gap. It even outperforms some early text LLMs of 7B like LLaMA-2-7B. Notably, our OpusLM-1.7B shows close performance with Moshi-7B (No.15) on MMLU (46.2 vs.49.8), even with 3x fewer parameters.
More importantly, we demonstrate that the text capability of the LLMs can be largely preserved during this speech-text joint pre-training. E.g., only 2.7 points of degradation on OpusLM-7B compared with OLMo2-7B.

\vspace{-7pt}
\subsection{Pre-Training Results} \label{exp:pretraining}
\vspace{-3pt}
\noindent \textbf{Scaling:} 
With the same training strategy, we compare the performance of OpusLMs of size 135M, 360M and 1.7B\footnote{We conduct this comparison before the annealing stage (\S\ref{sec:annealing})}. The results are shown in Fig.~\ref{fig:scaling}.
For speech capability, although the OpusLM-1.7B achieves impressive performance on both ASR and TTS tasks, significant degradation is observed on the 135M and 360M versions, especially the TTS-WER (38.7\% and 19.8\%, respectively).
Similarly, for the text capability, although SmolLM2-1.7B and OpusLM-1.7B achieve an MMLU of 51.7 and 46.2 respectively, all the corresponding 135M and 360M models get MMLU scores slightly higher than 25.0, which is nearly no better than a random guess\footnote{MMLU tasks are multiple choices among 4 options.}. 
The observed scaling property suggests the necessity of adopting a sufficiently large model (e.g., 1.7B) in this speech-text joint training setup.

\noindent \textbf{Design Choice Verification:}
We show the empirical evidence in Tab.~\ref{tab:pretraining} to support our findings of pre-training in \S\ref{training}. 
We find the effect of changing loss region at 250k updates is mitigated in our full-size 1.7B model training (Tab.~\ref{tab:pretraining}.a).
As in Tab.~\ref{tab:pretraining}.b, we tried to adopt the large batch size (i.e., 4M) like text LLM development \cite{llama, olmo20242}. With the same number of updates, we find the large batch size does not show a decisive advantage (especially on the MMLU score) even with 4 times more computing. 
Additionally, with our data amount, the 4M batch size is translated to 125k updates to the model, which is expected insufficient to help the model converge. So the batch size remains at 1M frames. 
%
Finally, in Tab.~\ref{tab:pretraining}.c, we show the importance of keeping the text corpus in-domain with the text LLM training corpus. Although the speech capability is not noticeably affected, the model's text capability would degrade considerably with improper text corpus adoption\footnote{Note the out-domain corpus in \cite{espnet_speechlm} is of sufficiently high quality, so we believe this degradation is more from domain mismatch.}.

\begin{table}[t]
    \centering
    \caption{Empirical Evidence of Pre-Training Design Choice}
    \vspace{-10pt}
    \scalebox{0.65}{
    \begin{tabular}{cccccc}
    \toprule
    \rowcolor{blue!10} \multicolumn{6}{l}{\texttt{a}. Impact of different loss region} \\ 
    \midrule
    \multicolumn{2}{c}{Loss Region} & \multicolumn{2}{c}{ASR (WER$\downarrow$)} & TTS (WER$\downarrow$) & Text \\ 
    \cmidrule(lr){1-2} \cmidrule(lr){3-4} \cmidrule(lr){5-5} \cmidrule(lr){6-6}   
        0-250k & 250-500k & T-Clean & T-Other  & T-Clean & MMLU ($\uparrow$)\\
    \hdashline 
        Whole  & \multirow{2}{*}{Target}  & \bf{2.8} & 7.2 & \bf{6.0} & 46.0 \\ 
        Target &  & 3.3 & \bf{6.9} & 6.2 & \bf{46.6} \\
    \midrule
    \rowcolor{red!10}\multicolumn{6}{l}{\texttt{b}. Impact of batch size} \\ 
    \midrule
    Batch Size & \#Updates & \multicolumn{2}{c}{Train Loss ($\downarrow$)} & Token Acc. ($\uparrow$) & MMLU ($\uparrow$) \\
    \hdashline 
    1M & \multirow{2}{*}{30k} & \multicolumn{2}{c}{2.369} & \textbf{0.162} & 45.7 \\
    4M &   & \multicolumn{2}{c}{{\textbf{2.322}}} & \textbf{0.162} & \textbf{48.3}\\
    \hdashline
    1M & \multirow{2}{*}{60k} & \multicolumn{2}{c}{2.254} & 0.178 & \textbf{47.2}  \\
    4M & & \multicolumn{2}{c}{{\textbf{2.235}}} & \textbf{0.181} &  46.5\\
     
    \midrule
    \rowcolor{orange!20}\multicolumn{6}{l}{\texttt{c}. Impact of in/out-domain text corpus} \\ 
    \midrule
     \multicolumn{2}{c}{\multirow{2}{*}{Text Corpus}} & \multicolumn{2}{c}{ASR (WER$\downarrow$)} & TTS (WER$\downarrow$) & Text \\ 
     \cmidrule(lr){3-4} \cmidrule(lr){5-5} \cmidrule(lr){6-6}   
     & & T-Clean & T-Other  & T-Clean & MMLU ($\uparrow$)\\
    \hdashline 
       \multicolumn{2}{c}{Text LLM Baseline}          &               &            &       & 50.7 \\ 
       \hdashline
       \multicolumn{2}{c}{In-Domain}         & \textbf{2.8}  & 7.2 & \textbf{6.0} & \textbf{46.0} \\ 
       \multicolumn{2}{c}{Out-Domain\cite{espnet_speechlm}} & \textbf{2.8}  & \textbf{5.9} & \textbf{6.0} & 30.5 \\
    \bottomrule
    \end{tabular}}
    \label{tab:pretraining}
    \vspace{-15pt}
\end{table}

\begin{table}[h]
    \centering
    \vspace{-5pt}
    \caption{Experimental Results on Annealing of OpusLM-1.7B. All results are in WER(\%$\downarrow$) except MMLU($\uparrow$). $\times$ means infeasible. * notes post-selection after 5 sampling. }
    \vspace{-10pt}
    \scalebox{0.7}{
    \begin{tabular}{l*{2}{>{\columncolor{blue!10}}c}*{2}{>{\columncolor{red!10}}c}*{2}{>{\columncolor{orange!20}}c}*{1}{>{\columncolor{yellow!20}}c}}
    \toprule
          & \multicolumn{2}{c}{\cellcolor{blue!10}In-Domain} & \multicolumn{2}{c}{\cellcolor{red!10}Out-Domain} & \multicolumn{2}{c}{\cellcolor{orange!20}Long-Form} & \multirow{1}{*}{\cellcolor{yellow!20}Text}\\
          \cmidrule(lr){2-3}\cmidrule(lr){4-5}\cmidrule(lr){6-7} \cmidrule(lr){8-8} 
          & ASR & TTS & ASR & TTS & ASR & TTS* & \multirow{1}{*}{\cellcolor{yellow!20}MMLU ($\uparrow$)} \\
    \hdashline
    Baseline              & 7.2 & 6.6 & 13.9 & 21.7 & $\times$ &$\times$ & 46.0 \\
    \ \ + Conti. Train    & 7.2 & 6.5 & 13.7 & 18.6 & $\times$ &$\times$ & 46.1 \\ 
    \ \ + Anneal (\texttt{Opt-A}) & \textbf{5.7} & 5.1 & \textbf{13.3} & 15.5 & 7.7 & 13.0 & \textbf{46.3} \\ 
    \ \ + Anneal (\texttt{Opt-B}) & 5.8 & \bf{4.0} & 14.5 & \textbf{12.7} & \textbf{4.7} & \textbf{11.5}  & 46.2  \\ 
    \bottomrule
    \end{tabular}}
    \label{tab:annealing}
    \vspace{-10pt}
\end{table}

\vspace{-7pt}
\subsection{Annealing Results} \label{sec:annealing}
\vspace{-3pt}
As described in \S\ref{training}, we demonstrate the benefit of annealing in Tab.~\ref{tab:annealing}. The experimental setups are in \S\ref{sec:setup}. Our observations are as follows:
(1) With both data compositions, we achieve improvement over the baseline and the continue training counterpart, which validates the benefit of annealing\footnote{The only exception is for out-of-domain ASR of \texttt{Opt-B}. Continue training setup means continuing the training without learning rate decay and data composition change for the same number of updates.}.
(2) The out-domain evaluation results imply the improvement is in general rather than adapting to the annealing data domain.
(3) Both long-form ASR and TTS become feasible after annealing, thanks to the inclusion of long-form training data. 
(4) Finally, although we see the benefits of annealing in both trials, the exact improvements can be significantly different (e.g., 15.5\% vs. 12.7\% in out-domain TTS). This observation suggests the importance of curating high-quality data composition during the annealing stage.


\vspace{-5pt}
\section{Conclusion}
\vspace{-5pt}
This paper presents OpusLM, a family of pre-trained speech language models that achieve competitive performance on speech recognition, speech synthesis, and text. The code, data, checkpoints, and training logs are released to support open speech language model research in the community.

\vspace{-5pt}
\section{Acknowledgement}
\vspace{-5pt}
Experiments of this work used the Bridges2 system at PSC and Delta system at NCSA through allocations CIS210014 and IRI120008P from the Advanced Cyberinfrastructure Coordination Ecosystem: Services \& Support (ACCESS) program, supported by National Science Foundation grants \#2138259,\#:2138286, \#:2138307, \#:2137603, and \#:2138296.

\begingroup
\tiny  
\bibliographystyle{IEEEtran}
\bibliography{mybib}
\endgroup

\end{document}